\documentclass[letterpaper, 10 pt, conference]{ieeeconf}  

\IEEEoverridecommandlockouts                              

\usepackage{amsmath,amsfonts}
\usepackage{algorithmic}
\usepackage{algorithm}
\usepackage{array}
\usepackage[caption=false,font=normalsize,labelfont=sf,textfont=sf]{subfig}
\usepackage{textcomp}
\usepackage{stfloats}
\usepackage{url}
\usepackage{verbatim}
\usepackage{graphicx}
\usepackage{xcolor}
\usepackage{hyperref}
\hypersetup{
    colorlinks=true,
    linkcolor=blue,
    filecolor=magenta,      
    urlcolor=cyan,
    pdftitle={Overleaf Example},
    pdfpagemode=FullScreen,
    }

\newcommand{\squeezeup}{\vspace{-2 mm}} 

\title{\LARGE \bf
Using Memory-Based Learning to Solve Tasks with State-Action Constraints
}

\author{Mrinal Verghese$^{1}$ and Christopher Atkeson$^{1}$
\thanks{This material is based upon work supported by the National Science Foundation under Grant IIS-1849287}
\thanks{$^{1}$Mrinal Verghese and Christopher Atkeson are with the Robotics Institute at Carnegie Mellon University {\tt\small mverghes, cga, @andrew.cmu.edu}}
}

\begin{document}
\maketitle
\begin{abstract}
Tasks where the set of possible actions depend discontinuously on the state pose a significant challenge for current reinforcement learning algorithms. For example, a locked door must be first unlocked, and then the handle turned before the door can be opened. The sequential nature of these tasks makes obtaining final rewards difficult, and transferring information between task variants using continuous learned values such as weights rather than discrete symbols can be inefficient. Our key insight is that agents that act and think symbolically are often more effective in dealing with these tasks. 
We propose a memory-based learning approach that leverages the symbolic nature of constraints and temporal ordering of actions in these tasks to quickly acquire and transfer high-level information. We evaluate the performance of memory-based learning on both real and simulated tasks with approximately discontinuous constraints between states and actions, and show our method learns to solve these tasks an order of magnitude faster than both model-based and model-free deep reinforcement learning methods.
\end{abstract}


\section{Introduction}
The family of tasks with constraints between states and actions has posed a significant challenge to many reinforcement learning-based algorithms. This family is comprised of tasks where certain affordances or actions may not be available or fail in some states due to mechanical constraints or other task constraints. Many tasks exhibit this structure including construction, assembly and disassembly, rearrangement, locking and unlocking, door operation, and certain navigation and other manipulation tasks. Even simple manipulation tasks like retrieving a mug from a cupboard require the cupboard to be open before the mug can be grasped. In each of these tasks, there may exist only a small set of action sequences that will actually solve the task. We will use the term {\it constraint task} to refer to tasks with constraints between states and actions.

In their work on the Montezuma's revenge problem, a constraint task, Ecoffet et al.\ highlighted the challenge constraint tasks pose to current reinforcement learning algorithms \cite{ecoffet_go-explore_2021}. We believe reinforcement learning algorithms struggle with constraint tasks for several reasons. The sequential nature of these tasks make it challenging to stumble on the reward, as the naive approach involves exploring all possible action sequences. Exploration time increases as a factorial of the total number of actions available in the task. 
In addition, neural network-based methods transfer information between versions of the task in an inefficient manner by storing that information in continuous learned weights rather than explicitly relating discrete high-level information. 
To overcome these challenges, we propose a memory-based learning agent that leverages the symbolic nature of the task constraints and temporal ordering of actions to efficiently explore the task and quickly transfer a learned task model to new instances.

Our contributions are as follows: 1) We design a dual controller for exploration and completion of constraint tasks. 2) We present a novel memory-based learning method to acquire a model of constraint interactions between components. This model can then be used by the dual controller to improve its performance. 3) We evaluate the performance of our method on a real mechanical locking task as well as a simulated disassembly task. 4)  We show our memory-based learner can be trained quickly on a small handful of task variations, and then generalize to unseen task variations.

\section{Related Work}
\subsubsection{Memory-Based Learning}
    In memory-based learning approaches all experiences are explicitly represented and stored in a memory. At test time, a relatively small subset of these experiences are indexed and a local model is fit to them. Memory based learning variants have been used for both robot control \cite{atkesonLocallyWeightedLearning1997a}\cite{ram_continuous_1997}\cite{schaal_memory-based_1994}\cite{atkeson_locally_1997}, and reinforcement learning \cite{martin_h_knn-td_2009}\cite{barreto_practical_2014}\cite{humphreys_large-scale_2022}. 
    Prior work has highlighted the advantages of memory-based learning including efficient learning, ease of adding new experiences to the agent by simply storing them in memory, the avoidance of catastrophic interference, and the effects of distribution shifts and long tailed distributions. However, memory-based approaches perform much more work at query time to find relevant memories. 
    

\subsubsection{Constraint Tasks}
    Many manipulation problems involve constraints and prior work has looked at solving these tasks. Mechanical locking tasks have been explicitly explored in prior work \cite{kulick_active_2015}\cite{baum_opening_2017}. Kulick et al.\ and Baum et al.\ formalized the exploration of joint dependency structures in puzzle box problems and tested multiple methods for physical exploration. We leverage their findings in our work as the exploration portion of our controller. More generally, robotic tasks can contain dependencies where the set of possible actions change discontinuously with with change in state. This can take the form of rearrangement planning \cite{krontiris_dealing_2015}\cite{krontiris_efficiently_2016}\cite{batra_rearrangement_nodate}\cite{gao_fast_2021}\cite{hartmann_long-horizon_2022}, navigation among movable obstacles (NAMO) \cite{levihn_autonomous_2014}\cite{sun_semantic_2019}\cite{wang_affordance-based_2021}, and manipulation among moveable obstacles (MAMO)  \cite{haustein_robot_2020}\cite{papallas_non-prehensile_2020}. Krontiris and Berkis, and Gao et al.\ presented the idea of using dependency graphs, where interactions between objects are represented in a graph structure for these types of problems \cite{krontiris_dealing_2015}\cite{krontiris_efficiently_2016}\cite{gao_fast_2021}. We leverage this idea of dependency graphs in our own work. One of the most common and well studied disassembly tasks, electronics disassembly, also features constraint structures \cite{bogue_robots_2019}\cite{vongbunyong_learning_2015}\cite{kristensen_towards_2019}\cite{foo_challenges_2022}\cite{du_learning_2022}. We include performance on a simulated electronics disassembly task to show the relevance of our method to common constraint tasks. 
    
    
\subsubsection{Symbolic Reasoning}
    Symbolic reasoning is often used to plan a series of high level actions in long horizon tasks using an abstracted task representation. Symbolic reasoning can leverage explicit \cite{fikes_strips_1971}\cite{nau_shop_1999} or learned \cite{konidaris_symbol_nodate}\cite{kaelbling_learning_2017}\cite{konidaris_skills_2018}\cite{huang_continuous_2019}\cite{garrett_online_2020}\cite{xu_deep_2021} abstractions. Huang et al.\ and  Konidaris et al.\ developed methods that reason about uncertainty over symbolic states when planning \cite{huang_continuous_2019}\cite{konidaris_skills_2018}. 
    Garrett et al.\ explore planning high-level actions over symbolic states where new observations can require a robot to update its belief about the world model and compute a new optimal series of actions \cite{garrett_online_2020}.

\section{Problem Definition}
\label{sec:Probelm_Def}

Constraint tasks contain a series of components or objects whose affordances might be limited by the state of the other objects in the environment. For a system with dynamics $x_{t+1} = f(x_t,u_t), x \in \mathcal{X}, u \in \mathcal{U}$, constrained systems are ones such that the subset of actions that will change the state of the task at any given time,  $\mathcal{U}_t' = \{u_t \in \mathcal{U} | f(x_t,u_t) \ne x_t\}$  is dependent on the state. Constraint tasks lend themselves well to symbolic representations as objects and the constraints between them are often discrete (there are no partial constraints). In these tasks an agent can usually discern whether an object's affordances are currently constrained through observation or attempted action. We can often further discretize the task by only considering component positions at their joint limits, turning a continuous state variable into a discrete one.
This however does not work for some constraints such as combination locks, where intermediate positions of degrees of freedom need to be attained. While these symbolic abstractions do not perfectly capture all nuances of constraint tasks, they enable the high-level reasoning needed to efficiently solve the task. Agents operating in a symbolic problem space should be robust to errors both in the symbolic state and actions, and in the relationship between symbols and continuous reality.

In our work, we consider mechanical locking puzzles as an example of constraint tasks. Locking puzzles contain a series of components that mechanically lock each other in a manner not immediately apparent to an agent. To solve the task the correct action sequence needs to be identified to unlock all components. In the locks we consider in this paper, each component has two positions defined by the joint limits of the component. In their prior work on these tasks, Kulick et al.\ \cite{kulick_active_2015} discuss the motivation of discretizing components into states at their joint limits for locking tasks. 
For a puzzle with $N$ components, we consider an $N$ dimensional state space where the state $x$ is part of the set $\{x|x \in [0,1]^N\}$. Each step, an agent can choose to try and toggle any of the $N$ components, so the action is an integer between 1 and N. Each of the $N$ components can lock any other component and cause an action on that component to fail. When the agent attempts to toggle any component, it will only successfully change its position if there are no other components locking it in their current states.
If no other components lock a component, that component is considered independent. A puzzle box is considered solved when its last component is moved to the 1 position.

\begin{figure}
    \centering
        \includegraphics[width=.34\linewidth,trim={40cm 0cm 10cm 0cm},clip]{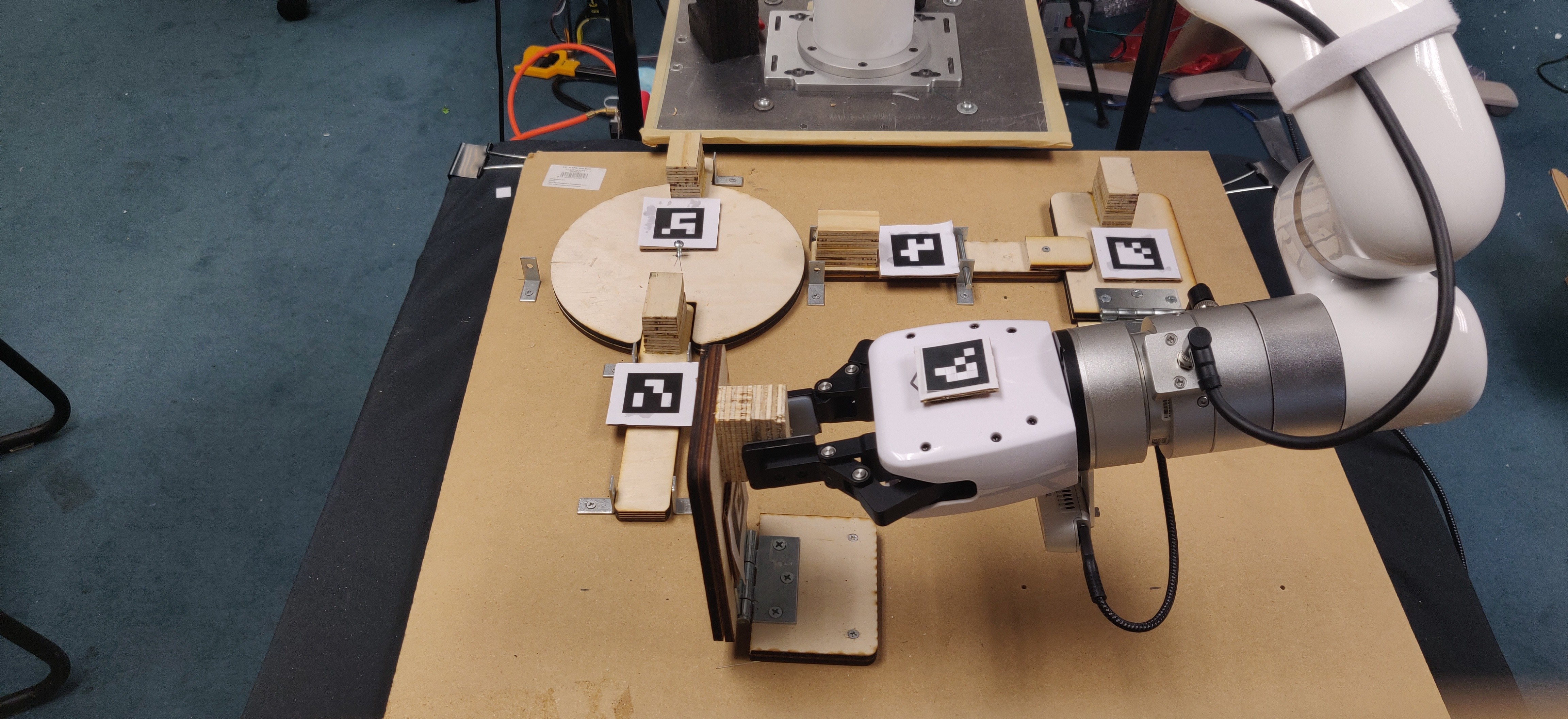}
       \includegraphics[width=.34\linewidth,trim={40cm 0cm 20cm 6.7cm},clip]{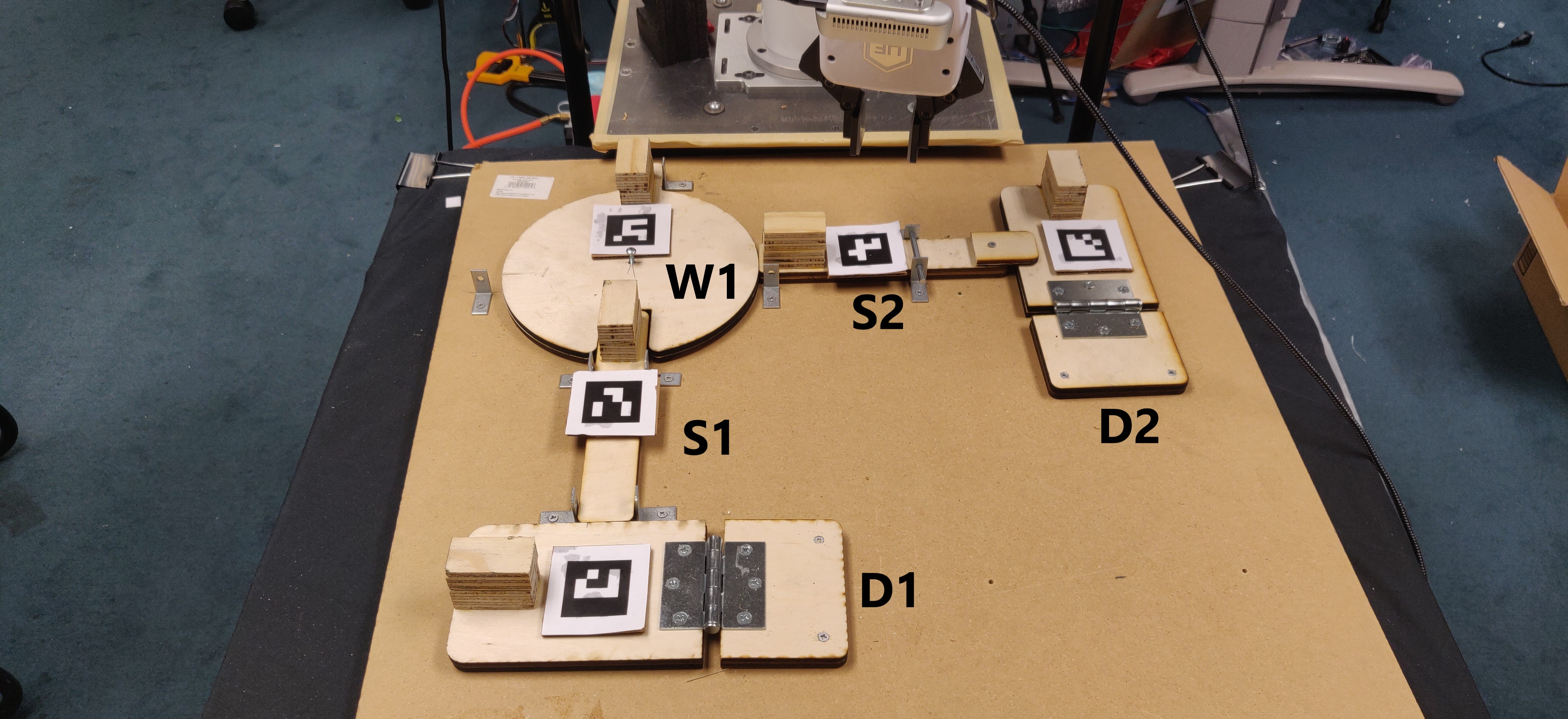}
       \includegraphics[width=.29\linewidth,trim={2cm 0cm 0cm 0cm},clip]{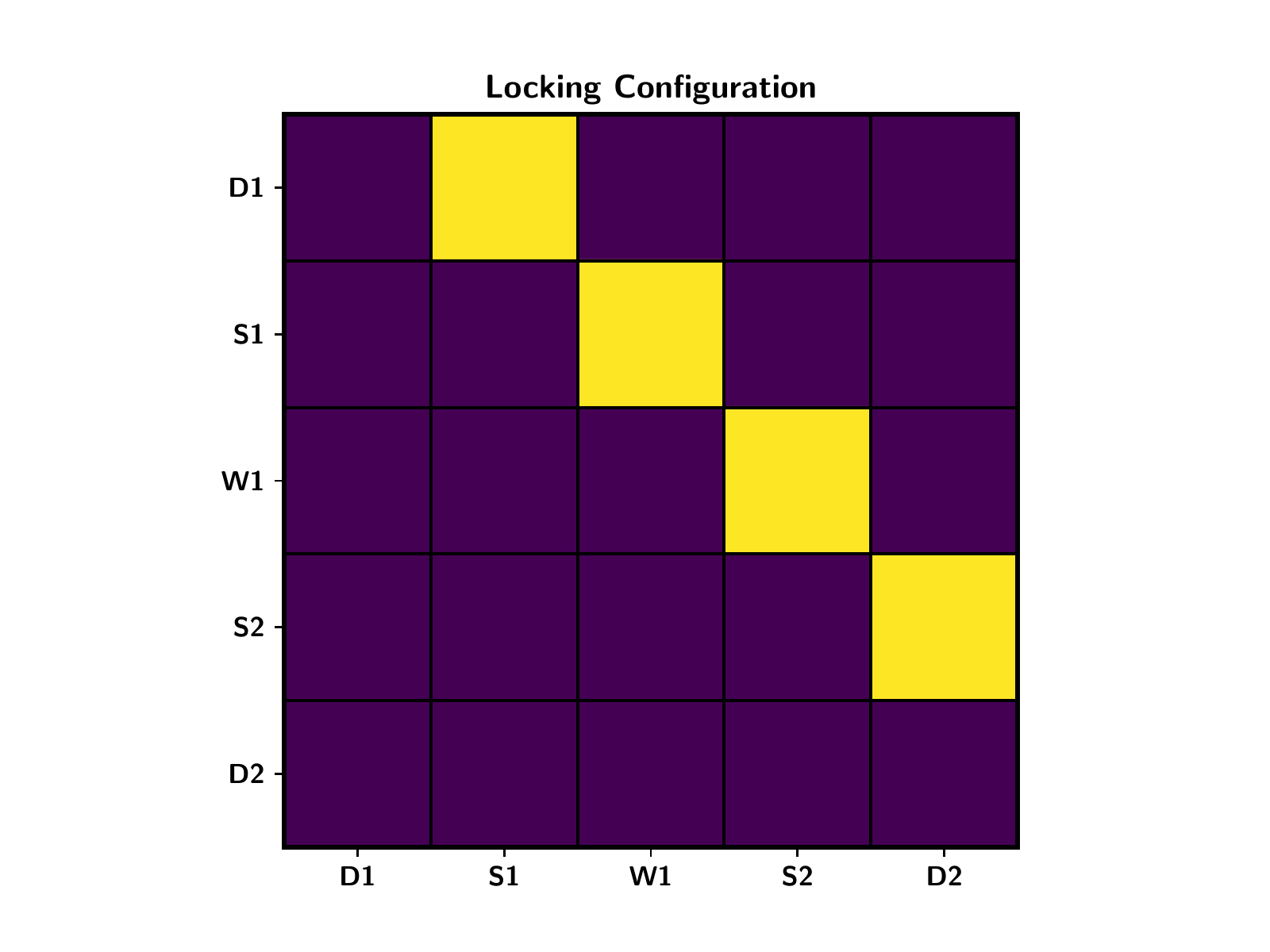}

     \caption{Left: Our robot solving a locking puzzle problem. Middle: One instance of the locking puzzle task used in this work. In order from bottom left to top right, the components are: door 1 (D1), slide 1 (S1), wheel 1 (W1), slide 2 (S2) and door 2 (D2). The components lock each-other in this order.
     The labels are randomly ordered when presented to the robot.
     Right: a symbolic representation of the locking configuration where cell $i,j$ indicates if component $i$ locks component $j$. A component with no locking relationship in its column is considered independent. The environment is considered solved when door 2 is opened.}
  \label{fig:Puzzle_Box_Closeup} 
  \vspace{-6mm}
\end{figure}

Following Baum et al.\ \cite{baum_opening_2017} we consider planar puzzle boxes comprised of three different components, doors, slides, and wheels. Doors are revolute joints that revolve out of the plane, slides are prismatic joints that slide within the plane, and wheels are revolute joins that revolve within the plane. Figure \ref{fig:Puzzle_Box_Closeup} shows an example puzzle task. In our locking puzzle tasks, doors and slides can lock each other based on placement, and slides and wheels can lock each other also based on placement. Wheels and doors however, do not interact. Slides have an overhanging piece to prevent doors from opening, and doors can prevent slides from traveling along their axis. Slides can also fit into a notch on a wheel to prevent it from rotating, and wheels can prevent a slide from traveling along its axis until the notch in the wheel is rotated to align with the slide. The locking puzzle task is an example of a manipulation task where components have underlying discontinuous interactions and learning these interactions can help agent quickly solve new task variations.

\section{Methods}
We design a dual controller that simultaneously explores and controls. Our controller is capable of learning the constraint structure of the problem online by constructing and maintaining a dependency graph to represent its beliefs. Our controller will eventually solve a given constraint task 
but its performance is poor as it begins each new task with no prior knowledge. We then add a memory-based learner to transfer knowledge between tasks and solve them faster. The memory-based learner stores the learned dependency graphs from every task it has previously seen. When faced with a new task, the agent picks the most relevant portions of prior dependency graphs, and uses them to initialize the new dependency graph, speeding up convergence of the controller.

\subsection{Dual Controller}

Dual control problems are a subset of adaptive optimal control problems where an agent may need to estimate information about its environment to solve a task. Dual control problems have been extensively studied \cite{feldbaum_dual_1963}\cite{unbehauen_control_2009}\cite{klenske_dual_2016} and have close ties to the exploration-exploitation problem in reinforcement learning. In constraint tasks, the agent should take actions that both identify the constraint structure of the problem, and make progress towards the goal. Dual controllers typically contain a utility function to estimate the costs and benefits of information gathering and taking actions. This allows dual controllers to automatically balance information gathering and control by optimizing this utility function, as well as avoiding unnecessary learning that does not reduce future costs. Specifying this utility function for a task can be difficult, so we create a heuristic dual controller with two explicit objectives: an exploration policy that seeks to identify the constraint structure of the task, and an exploitation policy that takes the actions with the highest expected probability of reaching the goal. Each of these policies output a value for each action in the action space, and an action is chosen based on an entropy-based weighting between the action values of the two policies.

\subsubsection{Exploration Policy}
We represent information about the constraint structure as a belief state over a dependency graph between components. Our dependency graph contains a node for each component. A directed edge exists in the graph if a component in a certain position constrains another component from actuating. The belief state represents how confident the agent is on the existence of each dependency graph edge. The exploration policy's goal is to confidently estimate the existence of each edge in as few actions as possible. This approach builds on work from Kulick et al.\ and Baum et al.\ \cite{kulick_active_2015}\cite{baum_opening_2017}.




To choose actions that effectively explore, we select actions that could maximally change our belief state, and thus provide the agent with the most information. By this definition, the optimal action is one that maximizes the expected KL-Divergence between the current belief state and the posterior belief state after that action is taken. This approach typically has lower costs than entropy minimization methods for exploration and allows the agent to recover from bad priors in its belief state \cite{kulick_advantage_2015}.

We represent our belief state as a set of random variables $E_t = \{P(e^{ij}) | e^{ij} \in G\}$ for the existence of edges in the dependency graph $G$. The action values for $u_t=n$, the action of trying to change the position of the $n$th component, are
\begin{equation}
\begin{aligned}
    &D = KLD\Big(P(E_t)||P(E_{t+1} | E_t, x_{t}, u_{t} = n, x_{t+1})\Big)\\
    &Q_{info}(x_t,u_t = n) = \mathbb{E}_{x_{t+1} \sim P(\cdot | E_t, x_{t}, u_{t} = n)}\big[D\big]
\end{aligned}
\end{equation}

where $P(x_{t+1} | E_t, x_{t}, u_{t} = n)$ is the probability of arriving at state $x_{t+1}$, $KLD(P||Q)$ is the KL-Divergence between two distributions $P$ and $Q$, and $P(E_{t+1})$ is the posterior belief state. 

We calculate this posterior belief state using Bayesian inference. We consider the likelihood edge $e^i$ exists in the dependency graph where $e^i$ represents a constraint relationship between two components. To reason about whether a component is current constrained, we consider the distribution across all incoming edges to a node as well as the possibility that the component associated with that node is independent (not constrained by other components). For components $n$ we term this distribution $Z_t^n$. To simplify this problem, we assume each component in each position is either constrained by one other component or independent. Considering multiple constraining components expands the belief space computation combinatorially and our experiments show this simplification is still able to model more complex constraint relationships. After trying an action, we update our distribution across $Z^n_t$ using the observation of whether component $n$ successfully moved.
We use this method to calculate both the expected KL-Divergence using the marginal probability of component $n$ moving as well as updating the belief state after observing an action.

\subsubsection{Exploitation Policy}
To calculate the exploitation policy, we model the constraint task as a task with continuous probabilities of possible deterministic constraints existing.
Our exploitation policy uses the current belief state to calculate the action with the highest probability of reaching the goal state after taking action $u_t = n$ at time $t$. 
We pose this as a graph search problem where nodes are symbolic task states, and edge values are transition probabilities given by $P(x_{t+1} | E_t, x_{t}, u_{t})$. These transition probabilities can be calculated as the marginal probability of a component successfully actuating in a given task state. We implement Dijkstra's algorithm to search this graph and terminate when we reach a success state for the task. For each action $u_t = n \in N$ we get a possible subsequent state $x_{t+1}$ as well as a probability of that action succeeding $P(x_{t+1} | E_t, x_{t}, u_{t})$. We then search the graph from $x_{t+1}$ to get a trajectory $(x_{t+1},x_{t+2},\dots,x_T$). We can now express our optimal action values as
\vspace{-2mm}
\begin{equation}
\begin{aligned}
    &V(x_{t+1}) = \prod_{\tau = t+1}^{T-1} \gamma P(x_{\tau + 1} | E_t, x_{\tau}, u_{\tau})\\
    &Q_{exploit}(x_t,u_t = n) = \gamma P(x_{t+1} | E_t, x_{t}, u_{t}) V(x_{t+1})
\end{aligned}
\end{equation}
\vspace{-2mm}

where $\gamma$ is a discount factor to penalize longer trajectories with equal probability of completion.

\subsubsection{Full Controller}
We use an entropy-based weighting scheme to weight the action values from the exploration and optimal policies. When the belief state entropy is high, we want to take exploratory actions to improve our confidence over the dependency graph. When the belief state entropy is low, this signals we are confident about the dependency graph, and following the exploitation policy will lead us to the goal. For belief state entropy $h$ we weight the action values from the two policies as:
\begin{equation}
\begin{aligned}
    Q(x_t,u_t = n) &= \frac{h}{h_{max}}(Q_{info}(x_t,u_t = n))\\
    &+ (1-\frac{h}{h_{max}})Q_{exploit}(x_t,u_t = n)
\end{aligned}
\end{equation}
where $h_{max}$ is an observed max entropy value that can also be tuned. Finally at each step, we choose the action with the highest weighted action value. In practice we find that these two objectives work well together. The information maximization policy can quickly learn the dependency graph improving the accuracy of the exploitation policy, and the exploitation policy guides exploration to states relevant to completing the task.

\subsection{Learned Priors}
The key contribution of our work is a memory-based learning method to learn priors on belief states that allow the controller to quickly solve unseen task variations. To utilize information from previous tasks for new tasks, we make two assumptions: interactions between component classes have similarities that persist across task variations and component interactions are dependent on the relative positions of the two components. These assumptions are based on our knowledge of mechanical constraint interactions and indicate that if we saw component $c_i$ constraining component $c_j$  at a given relative position in a previous task, then component $c_i'$ and $c_j'$ of the same classes as the original components and at a similar relative position are likely to interact the same way in the current task. Even if these assumptions aren't always true they are very helpful for directing the agents exploration. Both these features are required to learn good component interaction priors. An agent that only pays attention to position difference may learn physically close components interact, but will be unable to tell in which direction the constraint relationship goes. An agent that only considers component classes can learn constraint relationships between components, but will assume these relationships exist at a distance as well which does not hold for mechanical constraints. 

Memory-based learning allows us to learn good priors from only a handful of past experiences. After completing a task, we take each pair of components in the learned dependency graph and store the value of the edge between them in a bucket depending on the classes of the two components. For a problem with $C$ component classes, this gives us $C^2$ buckets. We also store the likelihood that a component is independent in a separate bucket. Each bucket contains a space that represents the difference in pose between the two components. For planar relationships, a point in this space looks like $(\Delta x,\Delta y, \Delta \sin(\theta), \Delta \cos(\theta))$. Each stored edge is indexed within its bucket by the pose difference between the two components it connects.

At test time, when faced with a new task variant, we initialize our belief state using the selected dependency graph edges from prior experiences. For each pair of components in the new task variant, we look up the bucket that corresponds to these two components. We use K-Nearest Neighbors in the pose difference space to index the most relevant prior edges. We then average the priors on these edges and initialize the edge in the new dependency graph with this value. This process can be repeated for every pair of edges in the new graph to construct a the new dependency graph. Finally, we add entropy to the graph by reducing certainty of dependency graph edges to model uncertainty and enable the controller to recover from bad priors. This new belief state over the dependency graph is used to initialize the controller for rapid convergence.

\section{Experiments and Discussion}
\label{sec:result}
\subsection{Baselines}
    We compare our agent to two deep reinforcement learning (RL) agents, a model-free RL agent trained with Deep Q Learning \cite{mnih_human-level_2015} and a model-based RL agent inspired by \cite{sharma_relational_2020}. The model-based agent attempts to learn a precondition model to predict which components are likely to be currently unlocked in any given state. We use this learned precondition model with the exploitation planner 
    from our dual controller to select the action with the highest probability of reaching the goal. The deep learning methods both leverage the same symbolic abstraction we give our memory-based method. Both the model-free and model based agents use Deep Networks with Graph Convolution Layers following the architecture by Battaglia et al.\ \cite{battaglia_relational_2018}. When training these agents, the numerical identifier, $1,\dots,N$, of each component was randomized during training to prevent agents from learning a static policy that opened components based on their numerical identifier. We also had to continually retrain agents on previous task variations to prevent catastrophic forgetting. We  add MT-OPT as a Multi-Task Reinforcement Learning baseline \cite{kalashnikov_mt-opt_2021}. MT-OPT adds a cross-task data sharing metric to training as well as balancing the number of successful and unsuccessful episodes in the training buffer.  
\squeezeup
\subsection{Simulation Results}
\subsubsection{Locking Puzzle}
\begin{figure*}
    \centering
       \includegraphics[width=.32\linewidth,trim={0cm 0cm 0cm 0cm},clip]{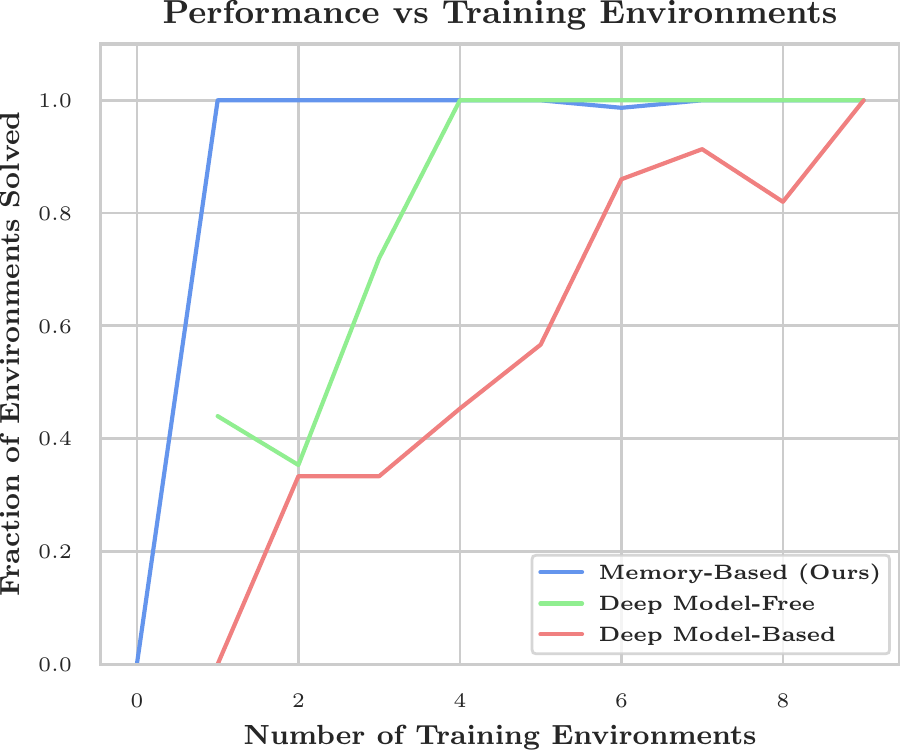}
       \includegraphics[width=.32\linewidth,trim={0cm 0 0cm 0cm},clip]{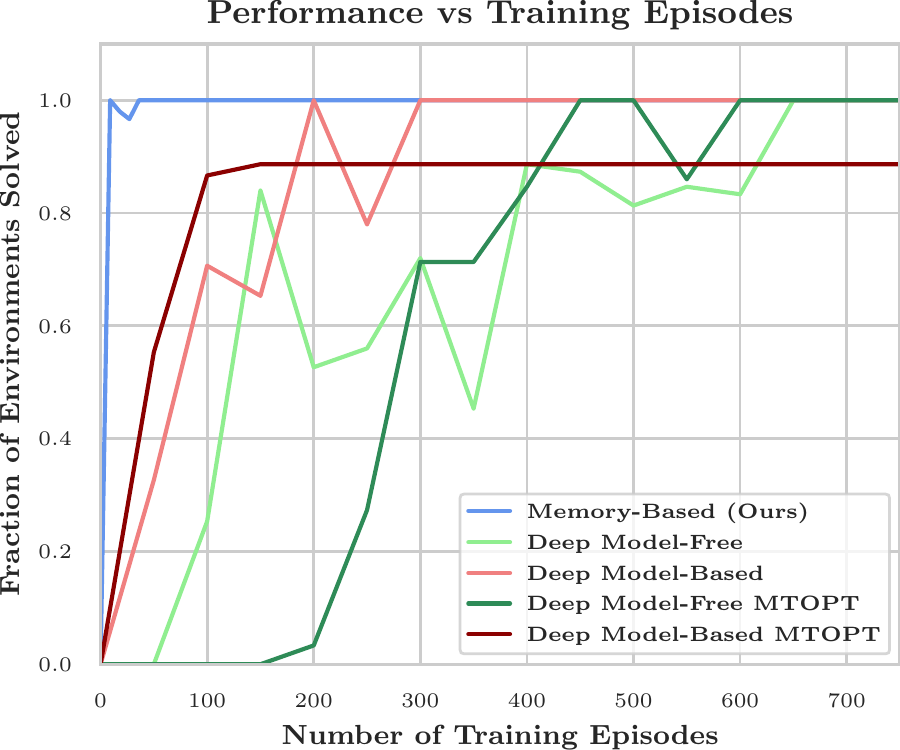}
       \includegraphics[width=.32\linewidth,trim={0cm 0 0cm 0cm},clip]{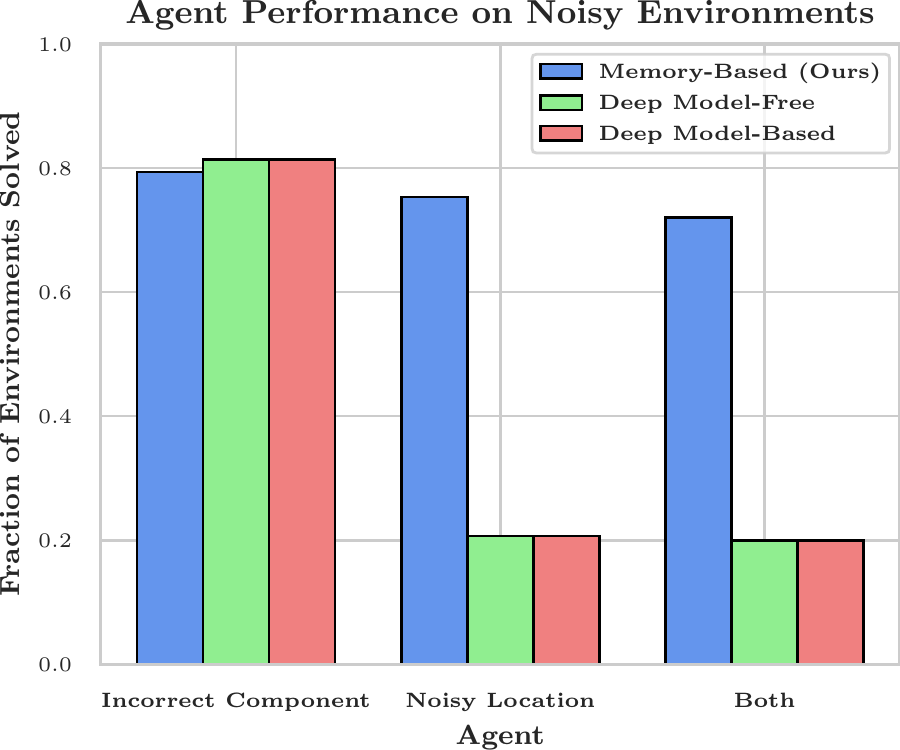}
     \caption{Left: Agent performance on unseen locking puzzle variants with respect to number of tasks variants trained on. Middle: Agent performance on unseen locking puzzle variants with respect to number of total training episodes. Agent performance on tasks with errors in the symbolic representation of the task. The three scenarios tested are, incorrect component identification, noisy component locations, and both these errors simultaneously. Performance is measured as the percent of tasks solved in 15 actions across 50 trials per test environment.}
  \label{fig:Env_Gen} 
  \vspace{-3mm}
\end{figure*}

    We first evaluate our method in a simulated symbolic environment. This simulated symbolic environment matches our symbolic description of locking puzzles in section \ref{sec:Probelm_Def} with puzzles comprising of 5 sequentially locking components. For each permutation of puzzle components, we procedurally generate multiple realistic layouts of these components that can be sampled. We test our memory-based learner against the Deep RL agents mentioned above. We provide the agents with up to 9 different training environments, each representing locking puzzle variants, with randomized component locations. We evaluate the performance of the agents on three unseen locking puzzle variants. Each of the test variants contains a new permutation of components never seen in the training set. Figure \ref{fig:Env_Gen} shows the performance of these agents on the set of unseen test puzzle boxes. Performance is defined as the percent of test tasks that are solved in under 15 actions.
    
In the left plot of figure \ref{fig:Env_Gen}, we varied the number of training task variants the agents had access to. All agents in this experiment were trained until convergence on the set of environments available to them. Our memory-based agent quickly acquires the task even with access to only one training environment. The Deep RL agents take longer with the model-based RL agent requiring training on all 9 training tasks to solve all test tasks. In the middle plot, we gave the agents access to all 9 training tasks but varied the total number of episodes they were allowed to train for. Here we included our Multi-Task RL baseline as well. The memory-based agent is able to generalize to unseen tasks quickly and solves every unseen environment after fewer than 30 episodes of training.  In addition, after seeing each environment only once, only 9 training episodes total, the memory-based agent is able to achieve very reasonable performance on unseen tasks. By contrast, the model-based and model-free agents take 300 and 650 episodes  respectively to solve the task, a full order of magnitude more than the memory-based agent. The Multi-Task RL baselines seem to provide slightly smoother convergence, but overall do not perform significantly better than the other baselines. Notably, the model-based Multi-Task baseline is never able to get more than 90\% success on the task. Compared to the deep learning baselines, the memory-based agent is able to learn tasks and transfer knowledge with access to fewer tasks and training episodes by storing prior experiences explicitly rather than in weights in a large network.

We also test the robustness of the agents in scenarios where the symbolic representation of the locking puzzle may be incorrect. We test three scenarios, one where a component has been incorrectly identified as a different type of component, one where Gaussian noise has been added to the component locations, and one where both of the above are true. Agents here have been trained until convergence on all 9 training task variations. The right figure shows the performance of the tested agents on the 3 unseen task variants with the aforementioned errors in the symbolic task representation. Performance of all agents drops a bit, but the Deep RL agents are particularly sensitive to noise in component location. By integrating a controller into our memory-based agent, we build a agent capable of recovering from errors by adjusting its model of the environment online.



\subsubsection{Disassembly Environment}

We also test our method on a more realistic environment. Electronics disassembly is another common task that contains a constraint structure. Many components cannot be removed until screws, panels, cables and other components have been removed first. In addition, this problem is well studied \cite{bogue_robots_2019}\cite{vongbunyong_learning_2015}\cite{kristensen_towards_2019}\cite{foo_challenges_2022}\cite{du_learning_2022} and multiple approaches have looked at symbolic decompositions of this task \cite{vongbunyong_learning_2015}\cite{du_learning_2022}. Prior approaches have often explicitly provided the component disassembly order to the agent, which is practical for disassembling a single type of electronic device, but can become tedious if an agent is to disassemble a wide range of devices. We test our method to see if it can learn the correct disassembly order without supervision.

To construct realistic simulations, we base this disassembly task on taking apart real computers in the lab. The goal of these tasks was to remove the motherboard from the case, but to accomplish that goal, the case screws had to first be removed, then the side panel, then the RAM and CPU as well as the GPU cables followed by the GPU, and then finally the motherboard could be removed. We assume access to a perception pipeline to locate components, skills or primitives to carry out the specific removal of each type of component, as well as a success indicator if the skill succeeded. We measured the 3D locations and orientations of each of the components for two lab computers, one to use for training, and one to test on. We selected 8 components from each computer to simulate, and the computers varied in number of RAM sticks and GPU cables, as well as the locations of every component.

This disassembly task provides an interesting challenge over our previous locking environment. The larger number of components, as well as many more component types, forces the agent to reason about many more interactions between components. In addition, unlike the locking task which had a entirely sequential locking structure, this computer disassembly task has many possible orders to remove motherboard components. Lastly, while the locking puzzle dealt with mostly planar components, the computer disassembly tasks deals with components with 6-DOF poses in 3D space.

\begin{figure}
    \centering
    \begin{minipage}[c]{0.5\linewidth}
    \includegraphics[width=\linewidth]{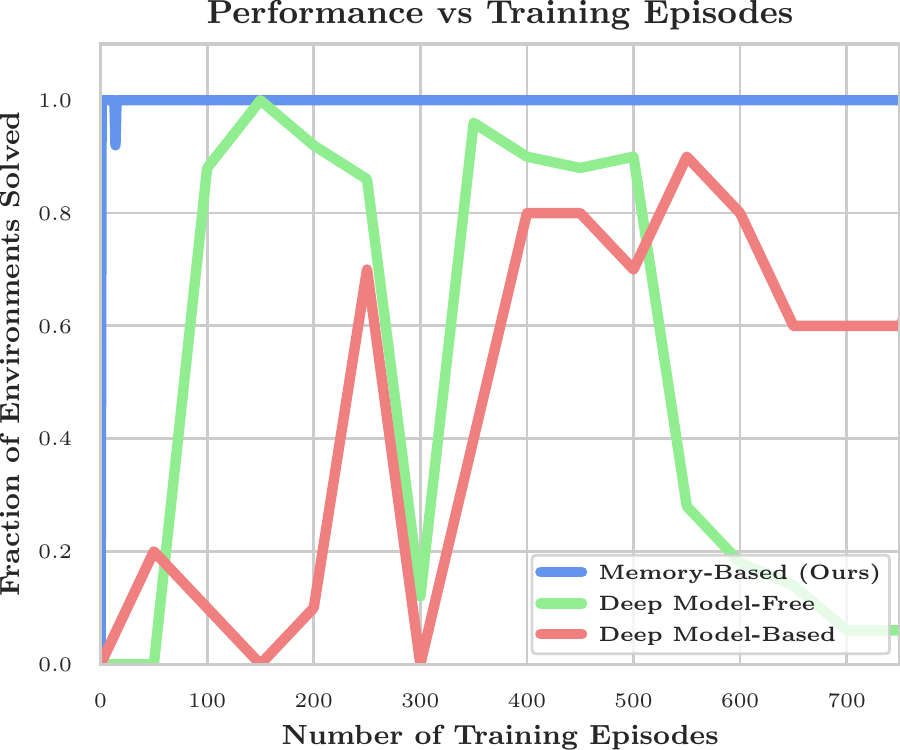}
  \end{minipage}\hfill
  \hspace{1mm}
  \begin{minipage}[c]{0.47\linewidth}
    \caption{Agent performance on an unseen computer disassembly environment with respect to number of total training episodes. Our memory-based method converged after 20 episodes.
     }
  \label{fig:Disassemb} 
  \end{minipage}
  \vspace{-3mm}
  \squeezeup

\end{figure}

We tested our deep model-free and deep model-based baselines in addition to our memory-based method on the computer disassembly task. As there was only one training environment, we elected not to test our Multi-Task RL baselines as they would provide no advantage over the regular baselines. Success in this task was defined as removing the motherboard within 24 actions. Figure \ref{fig:Disassemb} shows the agents' performance on an unseen test disassembly task.

After only one training episode, our memory-based agent was able to exploit the similarities between the constraint interactions of the training and test computers and solve the task in 20 actions. Further training helped reduce this to 17 actions and we stopped training after 20 episodes as we saw no further improvement. Both baseline agents had trouble reliably converging to a solution, and no amount of hyper-parameter tuning seemed to improve this. In addition, both deep learning agents required us to augment the environment with dense rewards for each component successfully removed, rather than the sparse reward the memory-based method got for completing the task. This could be attributed to the much more challenging nature of this disassembly task as described above. The model-free baseline was able to briefly reach 100\% success after 150 training episodes, a full order of magnitude slower than our memory-based agent. This task also breaks our assumption that each component is constrained by at most one other component. Nevertheless, our method is still able to effectively model constraints in this task and transfer them to the test environment. This experiment serves to show the performance of our method on more realistic tasks, and we hope to evaluate our method on a real-world disassembly task in the future.

\squeezeup
\subsection{Sim-to-Real Performance}

We investigate how the agents trained in simulation perform on the real locking puzzle tasks. To operate on a real robot, we use Aruco markers \cite{garrido-juradoAutomaticGenerationDetection2014a} to identify puzzle box components and the component's type and position. We design a set of behavioural primitives for actuation of revolute and prismatic joints. Given a joint axis, joint origin, and grasp point, these behavioural primitives are capable of actuating any revolute or prismatic joint between joint limits. We use an xArm-7 to manipulate components with a force torque sensor in its wrist to identify when components are locked and do not move. We hold the real world puzzle box out of the training set and train the agents to convergence on the other simulated training environments. We evaluate the average performance of the agents over 5 attempts to solve the puzzle box. Agents time out and the trial is stopped if they exceed 30 actions in a trial. Our memory-based agent is able to solve the real world environment in an average of 6.6 actions, the model-free agent averages 5.4 actions, and the model-based agent struggled averaging 21.2 actions (3 of its 5 trials timed out). The poor perfomance of the model-based agent is likely due to some fundamental dissimilarity between the procedurally generated environments, and our real puzzle boxes. The deep model-free method was able to slightly outperform our method, but this was after 700 training episodes as opposed to the 30 episodes our method needed. Please see the accompanying video (\url{https://youtu.be/EzPeMsq1g0M}) for footage of our method running on the real robot.

\squeezeup
\subsection{Limitations}
    It is important to note the limitations of our method. First and foremost, we assume access to both the vision and action components that allow us to operate in a symbolic space. Manually specifying these for each task can be cumbersome. Thankfully there is a large body of work on learning to ground symbols from vision and learning discrete behavioral primitives or skills \cite{huang_continuous_2019}\cite{wang_learning_2019}. Furthermore, we showed our agent is robust to errors in the  estimate of the task state. We currently assume each component is constrained by either one or zero other components. We would like to augment our dependency graph to explicitly model multiple constraint relationships for a single component. We also assume components can be classified into predefined classes. In the future we seek to break this assumption by utilizing distance in a visual embedding space or shape representations like voxels or point clouds to define similarity between new components and previously observed components. There are also certain limitations that stem from our symbolic assumptions. We assume our agents know how to discretize a components continuous action space into discrete values. We also assume we can always accurately detect when an action fails. Lastly, we assume our agents know ahead of time how each component in the problem should move. Breaking each of these assumptions is important for deployment of this method to a broad range of constraint tasks, and we will tackle these in future work.

\section{Conclusion}
\label{sec:conclusion}

	In this work, we presented an approach for symbolically representing constraint tasks and solving them with memory-based learning. Our agent is capable of learning with only a few episodes per task variation and can begin to generalize to unseen environments after only a couple of training tasks. This performance is enabled by indexing relevant prior experiences and using a controller that can adjust its model of object interaction online. We believe these ideas of leveraging symbolic structures and memory-based learning, show great potential for realizing the data efficient robot learning we want.
	
	Going forward, we would like to explore the use of ``pre-manipulation'', preliminary interaction with components, in addition to vision to better estimate relevant priors on constraint interactions. We would also like to consider problems with non-physical constraints. Lastly, we believe memory-based approaches can show promise in many other robot learning tasks outside constraint tasks, and we are eager to explore further.


\bibliographystyle{IEEEtran}
\bibliography{Lock_Problems}



 





\end{document}